\documentclass[11pt]{article}
\usepackage[final]{acl}

\usepackage{times}
\usepackage{latexsym}

\usepackage[T1]{fontenc}

\usepackage[utf8]{inputenc}

\usepackage{microtype}

\usepackage{inconsolata}

\usepackage{graphicx}

\usepackage{booktabs}       

\usepackage{framed}
\usepackage{listings}
\usepackage{xcolor}

\usepackage{multirow}
\usepackage{amsmath}
\usepackage{enumitem}
\usepackage{cleveref}
\usepackage{CJK}
\usepackage{tcolorbox}
\usepackage{balance}
\title{Do VLMs Read or Rewrite?\\[4pt]On Transcription Faithfulness in Vision-Language Models}

\author{Gwang Gook Lee, Kenan Emir Ak, Jay Mohta, Yan Xu, Dimitrios Dimitriadis  \\
        Amazon.com \\
        \texttt{\{gglee, kenanea, jaymoht, yanxuml, dbdim\}@amazon.com}
        }

\begin{document}
\maketitle

\begin{abstract}
Vision Language Models (VLMs) are increasingly used in place of traditional OCR pipelines for document understanding. In this paper, we show they do not always act as faithful transcribers: when text is imperfect, they often tend to rewrite it into a more plausible form -- a behavior that clean-text OCR benchmarks cannot detect. We introduce \textbf{FaithC4}, a multilingual perturbation benchmark of 1,455 single-page documents (English, Chinese, Korean) with three perturbation families: scramble, random substitution, and visually similar substitution. We use the benchmark to evaluate 15 systems spanning general-purpose VLMs, OCR-specialized VLMs, and traditional OCR pipelines. These three categories differ in WER degradation under perturbation: general-purpose VLMs degrade by up to 6.9 points, OCR-specialized VLMs by 0.1--3.4 points, and traditional OCR by less than 0.8 points on English. Probing Qwen3-VL-4B layer-by-layer, we identify a consistent pattern: rewriting fires only when a perturbed word's final layer FFN representation stays close to the original encoding; when the representation diverges sufficiently, the model transcribes faithfully. Word length affects rewriting rate: short words (4--6 characters) are rewritten up to 10\% of the time, with a sharp cutoff at 8 characters above which rewriting drops to 0\%.
\end{abstract}

\section{Introduction}
\label{sec:intro}

Vision Language Models (VLMs)~\cite{llava, internvl, qwenvl}, which pair a vision encoder with a Large Language Model (LLM), are increasingly used for document understanding. On standard Optical Character Recognition (OCR) benchmarks, VLMs often match or surpass traditional OCR pipelines~\cite{du2020pp, doctr2021, tesseract}, and a single VLM can handle text recognition, layout analysis, and downstream tasks such as document question answering within one model. As a result, practitioners are increasingly replacing traditional OCR pipelines with VLMs across a range of document processing applications.

However, this shift introduces a problem that standard OCR benchmarks do not capture. \textit{VLMs tend to rewrite text rather than transcribe it.}  When the input has a typo, a visual artifact, or an unclear character, the VLM often outputs what it expects to see instead of what is written. A VLM shown the typo \textit{``prbolem''}, for example, will often return \textit{``problem''}. This is reminiscent of the \textit{typoglycemia} effect in human reading, where readers recover meaning from scrambled letters \cite{Grainger2004}. Traditional OCR systems are not affected by this: they transcribe character by character and have no language model to override the visual evidence.

While the rewriting behavior can be beneficial for recovering some genuine errors, it may be problematic for some domains that may require literal transcription, such as legal documents, medical records, and historical manuscripts. Despite its importance, this rewriting behavior has been largely overlooked in VLM research. Existing OCR benchmarks~\cite{ocrbench,omnidocbench} focus on clean text recognition and are therefore insufficient for investigating how models can handle imperfect inputs.

In this work, we conduct a controlled  perturbation study to systematically investigate rewriting behavior in VLMs. Our contributions can be listed as follows:

\begin{itemize}[nosep]
    \item We evaluate 15 systems and find that perturbed text degrades transcription accuracy with the magnitude varying by system type: general-purpose VLMs (up to 6.9\,WER\,pp on English), OCR-specialized VLMs (0.1--3.4\,pp), and traditional OCR ($<$0.8\,pp).
    \item The findings hold across English, Chinese, and Korean, three typologically diverse scripts, though the most damaging perturbation type differs by script.
    \item We show that perturbations have non-local effects. Corrupting 5\% of characters drives errors on the unperturbed remainder up by as much as 7.3$\times$ vs.\ the clean baseline with general VLMs.
    \item Probing Qwen3-VL-4B layer-by-layer, we find that rewriting fires when the perturbed word's internal representation stays close to the original encoding. Word length also affects rewriting: short words (4--6 characters) are rewritten up to 10\% of the time, dropping to 0\% above 7 characters.
    \item We introduce \textbf{FaithC4}, a multilingual benchmark of 1,455 documents in English, Chinese, and Korean that measures transcription faithfulness under imperfect inputs, a property clean-text benchmarks cannot capture.\footnote{Dataset and evaluation code: \url{https://github.com/gwanglee/DoVLMsRead}}
\end{itemize}
\section{Related Work}
\label{sec:formatting}


\paragraph{Vision Language Models}


VLMs extend LLMs to handle visual inputs alongside text. A typical VLM consists of a pretrained vision encoder, a language backbone, and a connector that projects image features into the language embedding space~\cite{yin2024survey}. Recent open-weight families (LLaVA~\cite{llava}, InternVL~\cite{internvl,internvl35}, Qwen-VL~\cite{qwenvl,qwen3vl}, Gemma~\cite{gemma3}) and commercial systems (GPT-4V~\cite{gpt4}, Gemini~\cite{gemini}) now reach strong performance on visual reasoning, document understanding, and grounded generation. Despite these capabilities, the behavior of these models on imperfect text, has not been systematically investigated. We show that the same language priors that enable strong reading may also be introducing silent rewriting when met with imperfect text.

\paragraph{Optical Character Recognition}
OCR has evolved from traditional engineered pipelines like Tesseract~\cite{tesseract} and PaddleOCR~\cite{du2020pp} to deep learning approaches including CRNN~\cite{shi2016end} and transformer-based architectures~\cite{nougat}. More recently, OCR-specialized VLMs \cite{layoutlm,niu2025mineru25,paddle,deepseek2} combine visual encoders with language models, representing a leading approach for text-rich document processing. While existing benchmarks such as DocVQA~\cite{docvqa}, OCRBench~\cite{ocrbench,ocrbench2}, and OmniDocBench~\cite{omnidocbench} evaluate document understanding and text recognition, they are not designed to assess whether models faithfully transcribe text as written or silently correct them.

\paragraph{Robustness to Text Perturbations}
A line of NLP work studies how models behave on perturbed text. \citet{belinkov2017synthetic} showed that machine translation models are highly sensitive to character-level scrambles and noise. Subsequent studies extended this analysis to text classification~\cite{ebrahimi2018hotflip} and BERT-style encoders~\cite{pruthi2019combating}. More recently, \citet{cao2023unnatural} found that LLMs can recover scrambled words at near-human rates. We extend this perturbation methodology to the OCR setting, where the model receives the perturbed text as an image. Unlike text-only studies, where the model sees the scrambled tokens directly, our setting lets us measure whether the VLM rewrites what it is shown rather than transcribing it.

\paragraph{Hallucinations in Language Models}
Hallucinations, the generation of plausible but incorrect content, have been extensively studied in LLMs~\cite{maynez2020faithfulnessfactualityabstractivesummarization,Ji_2023}. In VLMs, recent work has focused mainly on object hallucinations, where models describe entities not present in the image~\cite{li2023evaluating, guan2024hallusionbench, wang2023amber}. Most VLM work examines adversarial text~\cite{shayegani2023surveyvulnerabilitieslargelanguage} and typographic attacks~\cite{wang2025textural}, bias~\cite{vo2025vision}, and the tendency of VLMs to ignore visual evidence in favor of language priors~\cite{liu2025seeing,lee2025vlind}. Closest to our setting, \citet{seeingbelieving} study OCR hallucination on document images whose answer fields are locally degraded by blur, occlusion, and low contrast. Their perturbation obscures characters without changing them, and their reference annotates each character by visibility, with fully illegible characters to be omitted rather than guessed; the property measured is thus whether a model recognizes that the visual evidence is gone. FaithC4 alters the characters instead of hiding them and renders the result at full legibility, so the evidence is always sufficient and the reference is exact. The two settings probe different failures: theirs, filling a gap the model cannot see; ours, overriding evidence the model can see because that evidence is improbable.


\section{Experiment Setup}
\label{sec:setup}

\subsection{Dataset}
\label{subsec:dataset}
Existing OCR benchmarks use pre-existing document images with fixed text, making targeted manipulation impossible. Instead, we start from raw text, apply controlled perturbations, and render the result as document images. The perturbed text is the ground truth, which lets us measure whether models faithfully transcribe what they see or silently ``rewrite" it.

\begin{figure*}[t!]
  \centering
  \includegraphics[width=0.9\textwidth]{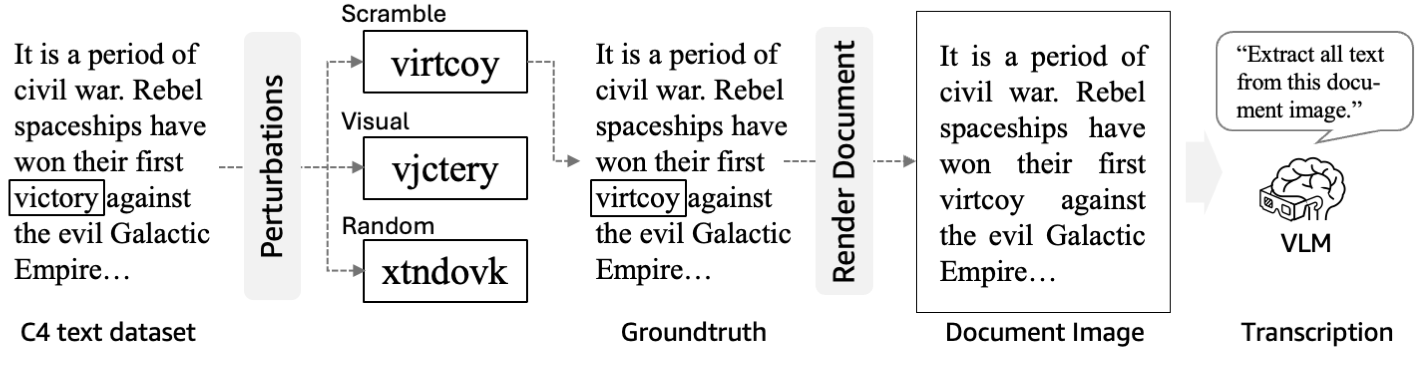}
   \caption{Dataset creation process. Source text from C4 is perturbed at the word level (scramble, visual, or random), producing modified ground truth. The perturbed text is rendered as a document image and passed to a VLM for transcription. Comparing the transcription against the perturbed ground truth reveals whether the model faithfully reads the modified text or ``rewrites'' it back to the original.}

   \label{fig:dataset-main}
\end{figure*}

We source text from the Colossal Clean Crawled Corpus (C4)~\cite{c4-raffel2020exploring}, a large-scale web-crawled dataset. From its multilingual subset (mC4), we sample documents in three diverse languages, namely English (EN), Chinese (ZH), and Korean (KO). For each language, we extract single-page documents by selecting text passages that fit within one page when rendered. This yields 1{,}455 documents (500 EN, 500 ZH, 455 KO), with averages of 599 words for English, 1{,}626 characters for Chinese, and 447 words for Korean.

\begin{table}[t]
\centering
\caption{C4-Long dataset statistics. Single-page dense PDFs (9pt, 1.2 line-height). Perturbation intensity calibrated to ${\sim}5\%$ of characters per document across all conditions and languages. Values are mean$\,\pm\,$std over documents.}
\label{tab:dataset}
\resizebox{\columnwidth}{!}{
\begin{tabular}{lccc}
\toprule
& \textbf{English} & \textbf{Chinese} & \textbf{Korean} \\
\midrule
Documents & 500 & 500 & 455 \\
Characters/doc & 3{,}531 $\pm$ 146 & 1{,}626 $\pm$ 135 & 2{,}214 $\pm$ 250 \\
Units/doc\textsuperscript{$\dagger$} & 599 $\pm$ 51 & 250 $\pm$ 69 & 447 $\pm$ 86 \\
\midrule
\multicolumn{4}{l}{\textit{Perturbation intensity (char-level change rate)}} \\
\quad Scramble & 5.15\% & --- & 4.81\% \\
\quad Random & 5.08\% & 5.07\% & 5.04\% \\
\quad Visual & 5.20\% & 5.03\% & 4.65\% \\
\midrule
\multicolumn{4}{l}{\textit{Units perturbed/doc (\% of units)}} \\
\quad Scramble & 7.3\% & --- & 6.9\% \\
\quad Random & 4.1\% & 9.8\% & 3.9\% \\
\quad Visual & 7.7\% & 18.4\% & 6.9\% \\
\bottomrule
\end{tabular}}
\begin{flushleft}
\footnotesize \textsuperscript{$\dagger$}Perturbation units: whitespace-delimited words (EN, KO) or non-overlapping 4-character CJK windows (ZH, since Chinese lacks word boundaries).
\end{flushleft}
\end{table}


We apply three types of character-level perturbations to eligible words (those with at least 4 characters), and keep the unmodified text as a control condition:

\begin{itemize}[nosep]
    \item \textbf{Original}: Unmodified documents serving as baseline.
    \item \textbf{Scramble}: Internal characters are shuffled while preserving the first and last (e.g., ``s\textit{tandar}d'' $\rightarrow$ ``s\textit{danart}d'').
    \item \textbf{Random}: Each character is replaced with a random letter from the same script (e.g., ``standard'' $\rightarrow$ ``\textit{xkpmewqi}'').
    \item \textbf{Visual}: Characters are replaced with visually similar alternatives (e.g., ``st\textit{a}nd\textit{a}rd'' $\rightarrow$ ``st\textit{s}nd\textit{o}rd''). List of visually similar characters are provided in Table~\ref{tab:visual_map}
\end{itemize}

Each perturbed text is rendered as a single-page PDF image (9pt serif font, 1.2$\times$ line-height). The generation process is shown in Figure~\ref{fig:dataset-main} and statistics are summarized in Table~\ref{tab:dataset}. Perturbation intensity is calibrated so that every language and condition changes ${\sim}5\%$ of the characters in a document (4.65--5.20\% across all eight language--condition pairs), which makes degradation comparable across scripts. Because eligibility differs by script, the fraction of \textit{units} perturbed varies to reach this common character-level intensity: 4.1--7.7\% of words in English.
In Chinese, which lacks word boundaries, we apply random and visual perturbations to non-overlapping 4-character windows (9.8--18.4\% of windows); scramble perturbation is omitted. In Korean, 3.9--6.9\% of words are perturbed. Because intensity is equalized at the character level, cross-lingual differences in degradation cannot be attributed to differing perturbation strength. See Appendix~\ref{sec:cjk_perturbation} for details.


\subsection{Models}
\label{subsec:models}

We evaluate 15 models spanning three categories:

\begin{itemize}
    \item \textbf{General-purpose VLMs}: open-weight (Qwen3.5-4B~\cite{team2026qwen3}, Qwen3-VL-4B~\cite{qwen3vl}, InternVL3.5-4B~\cite{internvl35}, Gemma4-E4B and -E2B~\cite{gemma42026google, gemma3}) and closed-weight
    (GPT-5.4-mini~\cite{singh2025openai}, Gemini-3-Flash, Gemini-2.5-Flash~\cite{gemini}) models.
  \item \textbf{OCR-specialized VLMs}: Models explicitly trained or fine-tuned for document text extraction (olmOCR-2-7B~\cite{poznanski2025olmocr}, DeepSeek-OCR-2~\cite{deepseek2}, MinerU2.5-Pro~\cite{niu2025mineru25}, PaddleOCR-VL-1.5~\cite{paddle}, LightOnOCR-2-1B \cite{lightonocr2025}).
    \item \textbf{Traditional OCR}: Character-level recognition pipelines (Tesseract~\cite{tesseract}, docTR~\cite{doctr2021}).

\end{itemize}

These categories differ in how much they rely on a language prior. General-purpose VLMs carry strong priors from broad text pretraining, OCR-specialized VLMs retain weaker priors after document-focused fine-tuning, and traditional OCR systems have essentially none.

\paragraph{Text Extraction.} We use a uniform extraction prompt across general VLMs:

\begin{tcolorbox}[colback=gray!5, colframe=gray!50, boxrule=0.5pt, left=4pt, right=4pt, top=2pt, bottom=2pt]
\small\textbf{Extraction Prompt:} ``Extract all text from this document image. Output document text only.''
\end{tcolorbox}

We provide no explicit instructions about handling perturbed text, allowing each model's natural behavior to surface. OCR-specialized models use their recommended prompts provided in Appendix~\ref{sec:app_models}.

\subsection{Evaluation Metrics}
\label{subsec:eval}

We evaluate OCR faithfulness with two complementary metrics, Word Error Rate (WER) for order-sensitive word-level accuracy and Edit Distance Similarity (EDS) for character-level similarity to the reference.

\paragraph{Word Error Rate (WER).} Our primary metric is the minimum edit distance between the reference and predicted token sequences, decomposed into substitutions ($S$), deletions ($D$), and insertions ($I$):
\begin{equation}
\text{WER} = \frac{S + D + I}{N}
\end{equation}
where $N$ is the number of tokens in the reference. Tokenization is whitespace-delimited for English and Korean. For Chinese, which lacks spaces between words, we measure at the character level, reducing WER to Character Error Rate (CER). WER is order-sensitive and captures extra tokens (insertions), missing tokens (deletions), and misreadings (substitutions).

\paragraph{Edit Distance Similarity (EDS).} Character-level normalized Levenshtein similarity:
\begin{equation}
\text{EDS} = 1 - \frac{\text{Levenshtein}(r, p)}{\max(|r|, |p|)}
\end{equation}
where $r$ and $p$ are the reference and prediction strings. EDS provides a character-granular complement to word-level WER, useful for detecting partial-word errors and near-miss transcriptions.


\paragraph{Normalization.} Before computing metrics, we normalize both reference and prediction. We flatten the text, remove special characters and non-textual artifacts, and collapse whitespace. Model-specific postprocessing further strips output tokens unrelated to document content; for example, DeepSeek wraps output in bounding-box tokens \texttt{<|ref|>...<|det|>[[coords]]<|/det|>}, which we remove before scoring.

\begin{figure}[h!]
\centering
\includegraphics[width=\columnwidth]{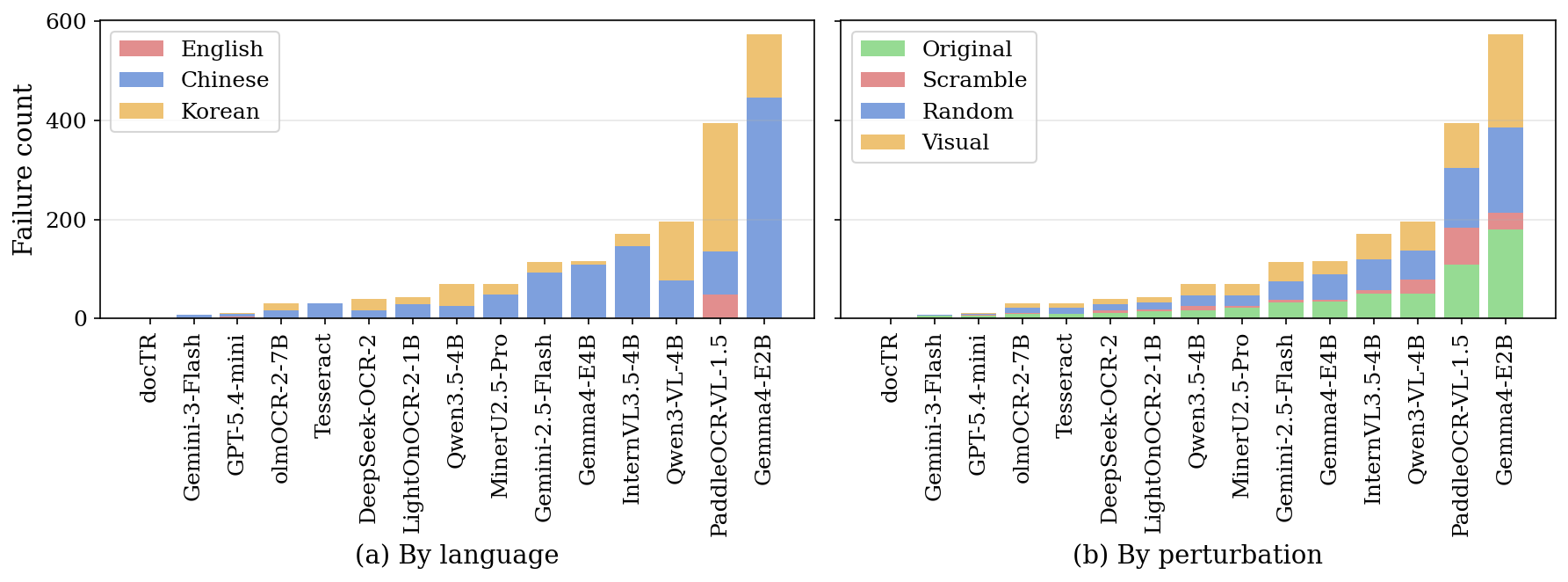}
\caption{Model failure counts by (a) language and (b) perturbation.}
\label{fig:doc-failure}
\end{figure}

\paragraph{Excluding Model Failures.} For each \textit{(model, language)} pair, we compute the document-level metrics of \S\ref{subsec:degradation} only on documents that the model processed successfully across all four perturbation conditions; the word-level analyses (Tables~\ref{tab:error_types} and~\ref{tab:collateral}) instead use all documents the model processed successfully under the relevant condition, which for most models is the same set to within two documents. Without this filter, performance deltas would conflate perturbation sensitivity with differences in which documents each model succeeded on, since failure rates vary by model. Figure~\ref{fig:doc-failure} provides the failure rate for each model and perturbation type. Model failure cases  such as refusals, repetition loops and script mismatches are analyzed in Appendix~\ref{app:failures}.

\section{Experimental Results}
\label{sec:results}

We evaluate 15 systems across English, Chinese, and Korean under the three perturbation types from \S\ref{subsec:dataset}. We measure performance degradation (\S\ref{subsec:degradation}) and probe Qwen3-VL-4B layer by layer to localize where rewriting originates (\S\ref{subsec:probes}).
\subsection{Performance Degradation and Error Analysis}
\label{subsec:degradation}


\begin{table}[t]
\centering
\caption{WER and EDS degradation in English dataset.}
\label{tab:degradation_en}
\resizebox{0.95\columnwidth}{!}{
\begin{tabular}{lrrrrrrrr}
\toprule
& & \multicolumn{3}{c}{$\Delta$WER (pp$\downarrow$)} & & \multicolumn{3}{c}{$\Delta$EDS (pp$\uparrow$)} \\
\cmidrule(lr){3-5} \cmidrule(lr){7-9}
\textbf{Model} & $n$ & Scr & Vis & Ran & & Scr & Vis & Ran \\
\midrule
Gemini-3-Flash & 456 & +0.13 & +0.23 & -0.01 &  & -0.01 & -0.03 & +0.00 \\
docTR & 500 & +0.12 & +0.45 & +0.11 &  & -0.02 & -0.09 & -0.03 \\
Tesseract & 500 & +0.17 & +0.72 & +0.20 &  & -0.03 & -0.13 & -0.04 \\
MinerU2.5-Pro & 500 & +0.61 & +0.84 & +0.10 &  & -0.17 & -0.14 & +0.01 \\
LightOnOCR-2-1B & 500 & +1.40 & +1.10 & +0.21 &  & -0.59 & -0.24 & -0.02 \\
Gemini-2.5-Flash & 451 & +0.97 & +1.65 & +0.68 &  & -0.36 & -0.51 & -0.32 \\
DeepSeek-OCR-2 & 499 & +2.47 & +1.39 & +0.53 &  & -0.74 & -0.30 & -0.13 \\
GPT-5.4-mini & 498 & +1.94 & +1.98 & +0.59 &  & -0.54 & -0.40 & -0.12 \\
Qwen3.5-4B & 498 & +2.29 & +1.95 & +0.81 &  & -0.71 & -0.41 & -0.26 \\
Gemma4-E4B & 500 & +2.44 & +2.23 & +0.79 &  & -0.83 & -0.57 & -0.17 \\
PaddleOCR-VL-1.5 & 466 & +3.19 & +1.70 & +0.64 &  & -2.09 & -0.66 & -0.23 \\
olmOCR-2-7B & 500 & +3.42 & +2.48 & +0.76 &  & -1.47 & -0.70 & -0.19 \\
Qwen3-VL-4B & 499 & +3.31 & +3.04 & +1.33 &  & -1.02 & -0.63 & -0.20 \\
Gemma4-E2B & 499 & +3.44 & +3.13 & +1.13 &  & -1.12 & -0.80 & -0.27 \\
InternVL3.5-4B & 498 & +6.90 & +6.77 & +3.21 &  & -2.99 & -2.26 & -1.03 \\
\bottomrule
\end{tabular}}
\end{table}

\begin{table}[t]
\centering
\caption{CER and EDS degradation in Chinese dataset.}
\label{tab:degradation_zh}
\resizebox{0.9\columnwidth}{!}{
\begin{tabular}{lrrrrrrr}
\toprule
& & \multicolumn{2}{c}{$\Delta$CER (pp$\downarrow$)} & & \multicolumn{2}{c}{$\Delta$EDS (pp$\uparrow$)} \\
\cmidrule(lr){3-4} \cmidrule(lr){6-7}
\textbf{Model} & $n$ & Vis & Ran & & Vis & Ran \\
\midrule
Qwen3-VL-4B & 458 & -0.07 & -0.22 &  & -0.11 & -0.03 \\
olmOCR-2-7B & 494 & +0.73 & -0.25 &  & -0.53 & +0.17 \\
MinerU2.5-Pro & 477 & +0.32 & +0.18 &  & -0.17 & +0.16 \\
PaddleOCR-VL-1.5 & 447 & -0.51 & +1.36 &  & -0.15 & -0.87 \\
Gemini-3-Flash & 493 & +0.57 & +0.37 &  & -0.56 & -0.40 \\
Qwen3.5-4B & 483 & +0.40 & +0.56 &  & -0.48 & -0.58 \\
Tesseract & 488 & +0.32 & +0.70 &  & -0.28 & -0.54 \\
DeepSeek-OCR-2 & 490 & +0.94 & +1.41 &  & -0.71 & -0.84 \\
InternVL3.5-4B & 407 & +1.87 & +1.51 &  & -1.98 & -1.73 \\
GPT-5.4-mini & 495 & +2.32 & +2.26 &  & -2.10 & -2.01 \\
Gemini-2.5-Flash & 420 & +2.69 & +2.52 &  & -2.29 & -2.18 \\
LightOnOCR-2-1B & 483 & +2.30 & +3.46 &  & -2.06 & -2.99 \\
Gemma4-E4B & 419 & +4.95 & +6.12 &  & -3.58 & -4.55 \\
Gemma4-E2B & 255 & +7.93 & +8.63 &  & -4.53 & -5.04 \\
\bottomrule
\end{tabular}}
\end{table}

\begin{table}[t]
\centering
\caption{WER and EDS degradation in Korean dataset.}
\label{tab:degradation_kr}
\resizebox{\columnwidth}{!}{
\begin{tabular}{lrrrrrrrr}
\toprule
& & \multicolumn{3}{c}{$\Delta$WER (pp$\downarrow$)} & & \multicolumn{3}{c}{$\Delta$EDS (pp$\uparrow$)} \\
\cmidrule(lr){3-5} \cmidrule(lr){7-9}
\textbf{Model} & $n$ & Scr & Vis & Ran & & Scr & Vis & Ran \\
\midrule
Gemini-3-Flash & 422 & +1.30 & +2.11 & +1.41 &  & -0.41 & -0.54 & -0.33 \\
Tesseract & 455 & +0.16 & +1.12 & +3.59 &  & -0.04 & -0.59 & -1.89 \\
MinerU2.5-Pro & 444 & +0.99 & +2.12 & +2.20 &  & -0.50 & -0.55 & -0.97 \\
PaddleOCR-VL-1.5 & 341 & +2.69 & +2.11 & +0.65 &  & -1.50 & -0.62 & +0.21 \\
DeepSeek-OCR-2 & 441 & +1.82 & +2.31 & +1.40 &  & -0.77 & -1.47 & -1.61 \\
LightOnOCR-2-1B & 450 & +1.90 & +3.21 & +2.86 &  & -0.58 & -1.12 & -1.57 \\
Qwen3.5-4B & 437 & +3.19 & +3.56 & +2.74 &  & -1.37 & -1.28 & -1.26 \\
Gemma4-E2B & 383 & +4.45 & +3.33 & +1.80 &  & -1.95 & -1.66 & -1.66 \\
Qwen3-VL-4B & 408 & +3.23 & +4.13 & +2.93 &  & -1.01 & -1.37 & -1.12 \\
Gemini-2.5-Flash & 406 & +3.92 & +4.06 & +2.62 &  & -1.86 & -1.84 & -1.67 \\
InternVL3.5-4B & 444 & +3.74 & +3.84 & +3.07 &  & -1.40 & -1.43 & -1.60 \\
olmOCR-2-7B & 447 & +3.41 & +3.96 & +3.36 &  & -1.43 & -1.60 & -1.91 \\
Gemma4-E4B & 451 & +4.26 & +4.45 & +2.63 &  & -1.97 & -2.20 & -1.76 \\
GPT-5.4-mini & 453 & +4.34 & +5.10 & +3.09 &  & -1.65 & -2.09 & -1.83 \\
\bottomrule
\end{tabular}}
\end{table}


Tables~\ref{tab:degradation_en}-~\ref{tab:degradation_kr} present performance degradation under perturbations across three languages. We report the change in percentage points relative to each model's baseline (original) performance. Absolute baselines are provided in Appendix~\ref{appendix:baseline}.

\begin{figure}[h!]
\centering
\includegraphics[width=\columnwidth]{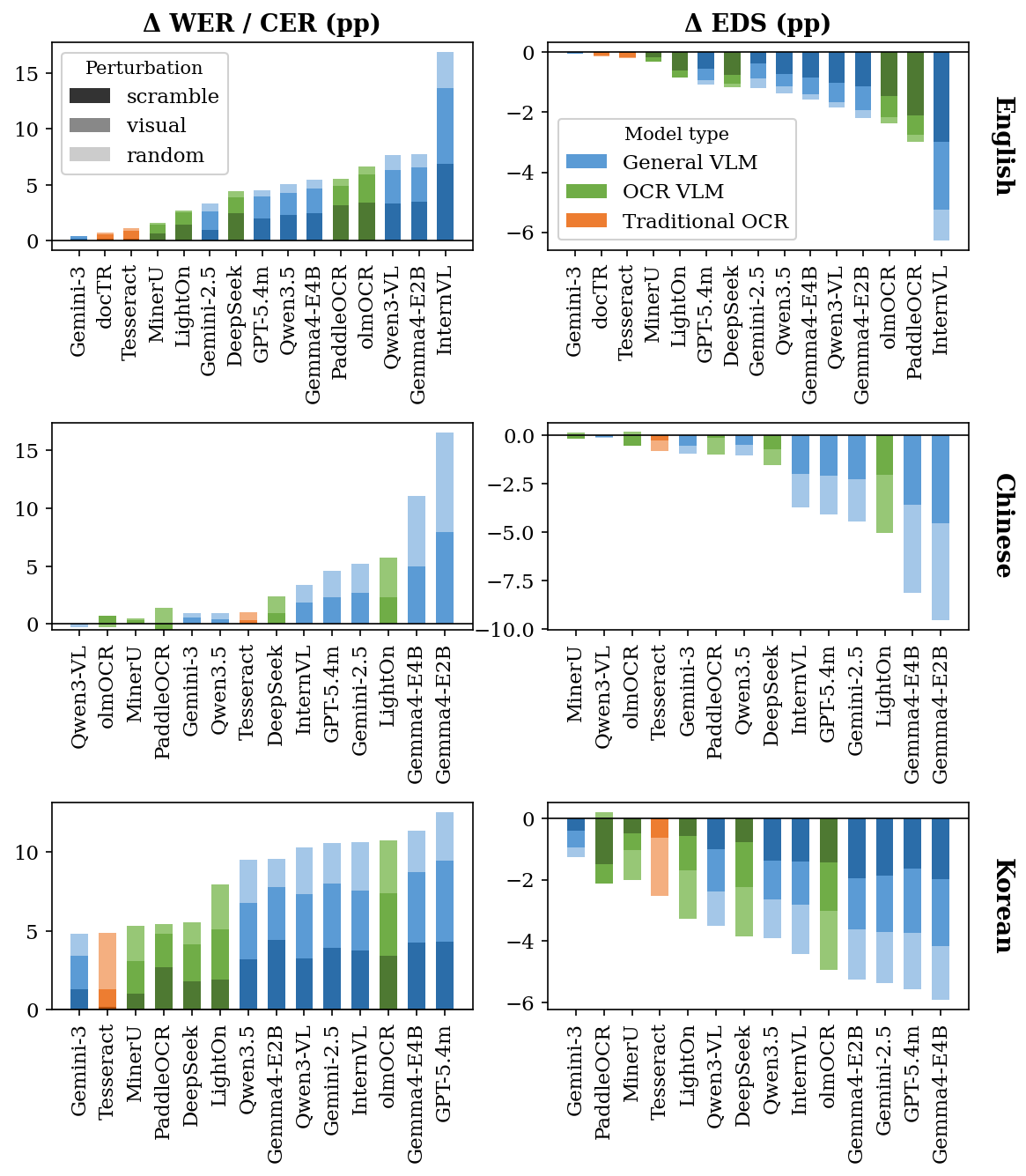}
\caption{Degradation ($\Delta$pp) by model, language, and perturbation type: WER/CER (left) and EDS (right), for English, Chinese, and Korean (top to bottom). Bars are \emph{stacked} over the three perturbation types, so total bar height is the sum of scramble, visual, and random rather than a single-condition value. Chinese has no scramble condition, and docTR is English-only.}
\label{fig:main-plot}
\end{figure}

\paragraph{Overall degradation.} Models exhibit three categories of robustness, as shown in Figure~\ref{fig:main-plot}; the ranges below are English $\Delta$WER across the three perturbations (Table~\ref{tab:degradation_en}). \textit{Traditional OCR} systems (Tesseract, docTR) are the most robust, degrading by at most 0.72pp, since character-level recognition is largely unaffected by within-word manipulations. \textit{OCR-specialized VLMs} span a wide range. MinerU (0.10--0.84pp) and LightOn (0.21--1.40pp) remain robust, while DeepSeek (0.53--2.47pp), PaddleOCR (0.64--3.19pp) and olmOCR (0.76--3.42pp) are markedly more sensitive. \textit{General-purpose VLMs} suffer the biggest degradation, with most models in the 0.59--3.44pp range and InternVL3.5-4B reaching 6.90pp. Gemini-3-Flash is the exception, degrading by at most 0.23pp, on par with traditional OCR. This ordering, traditional OCR $<$ OCR-specialized VLMs $<$ general-purpose VLMs, tracks the strength of each model's language prior, with stronger priors producing more rewriting behavior.


\paragraph{Language prior as error mechanism.}
\begin{table}[t]
\centering
\caption{Error type distribution on perturbed words under scramble (English). Each row sums to 100\%. Sorted by Rewritten rate.}
\label{tab:error_types}
\resizebox{\columnwidth}{!}{
\begin{tabular}{lcccc}
\toprule
\textbf{Model} & \textbf{Rewritten} & \textbf{Close} & \textbf{Moderate} & \textbf{Wrong} \\
\midrule
Tesseract & 0.00\% & 77.30\% & 12.06\% & 10.64\% \\
docTR & 0.00\% & 60.00\% & 14.19\% & 25.81\% \\
Gemini-3-Flash & 0.00\% & 50.00\% & 29.17\% & 20.83\% \\
Gemini-2.5-Flash & 1.65\% & 44.65\% & 29.16\% & 24.55\% \\
PaddleOCR & 7.87\% & 45.47\% & 6.89\% & 39.76\% \\
DeepSeek & 13.52\% & 45.41\% & 23.23\% & 17.84\% \\
MinerU & 16.16\% & 57.64\% & 9.17\% & 17.03\% \\
GPT-5.4-mini & 23.99\% & 63.16\% & 8.20\% & 4.64\% \\
Qwen3.5 & 27.35\% & 52.22\% & 9.56\% & 10.87\% \\
Gemma4-E4B & 30.25\% & 59.50\% & 6.50\% & 3.75\% \\
Qwen3-VL & 31.33\% & 44.46\% & 11.33\% & 12.89\% \\
Gemma4-E2B & 35.45\% & 51.74\% & 8.71\% & 4.09\% \\
InternVL & 46.37\% & 30.96\% & 14.80\% & 7.87\% \\
LightOn & 53.69\% & 29.80\% & 6.16\% & 10.34\% \\
olmOCR & 58.63\% & 32.09\% & 4.84\% & 4.44\% \\
\bottomrule
\end{tabular}}
\end{table}


To understand why general-purpose VLMs degrade more, we classify word-level errors at perturbed positions into four categories that separate rewriting from ordinary OCR mistakes. \textit{Rewritten} predictions match the original unperturbed word, the strongest evidence that the model has overridden the image with its language prior. \textit{Close} predictions fall within edit distance 2 of the perturbed text, the regime of single-character substitutions, deletions, or transpositions that any OCR system produces on noisy input. \textit{Moderate} predictions share at least half their characters with the perturbed text (edit-distance ratio $\geq 0.5$) but are not close enough to count as ordinary noise. \textit{Wrong} predictions are unrelated to either original or perturbed form. Table~\ref{tab:error_types} shows the distribution under scramble perturbation. Traditional OCR rewrites 0\% of perturbed words, while most VLMs rewrite a substantial fraction, confirming that language priors drive rewriting behavior. Rewrite rates vary widely across VLMs, from 0.00\% for Gemini-3-Flash and 1.65\% for Gemini-2.5-Flash to 27.35\% for Qwen3.5-4B and 58.63\% for olmOCR-2-7B.

\paragraph{Error propagation beyond perturbed words.}
\begin{table}[t]
\centering
\caption{Per-word error rate (\%) on perturbed vs non-perturbed words (English). Perturbing ${\sim}5\%$ of words causes collateral errors on unperturbed words too.}
\label{tab:collateral}
\resizebox{\columnwidth}{!}{
\begin{tabular}{l|c|cc|cc|cc}
\toprule
& \textbf{Orig} & \multicolumn{2}{c|}{\textbf{Scramble}} & \multicolumn{2}{c|}{\textbf{Random}} & \multicolumn{2}{c}{\textbf{Visual}} \\
\textbf{Model} & All & Pert & Non-pert & Pert & Non-pert & Pert & Non-pert \\
\midrule
Qwen3.5 & 0.35 & 3.00 & 2.18 & 1.35 & 0.89 & 3.03 & 2.01 \\
Gemini-3-Flash & 0.35 & 0.60 & 0.41 & 0.49 & 0.34 & 0.78 & 0.52 \\
LightOn & 0.36 & 1.96 & 1.66 & 0.55 & 0.48 & 1.79 & 1.30 \\
MinerU & 0.37 & 1.15 & 0.93 & 0.59 & 0.42 & 1.54 & 1.10 \\
Tesseract & 0.40 & 0.69 & 0.57 & 0.73 & 0.60 & 1.57 & 1.09 \\
Qwen3-VL & 0.40 & 3.95 & 2.93 & 1.77 & 1.39 & 4.30 & 2.91 \\
olmOCR & 0.42 & 4.81 & 3.76 & 1.41 & 1.11 & 3.88 & 2.72 \\
docTR & 0.57 & 0.80 & 0.68 & 0.81 & 0.69 & 1.41 & 0.98 \\
Gemma4-E4B & 0.68 & 3.92 & 2.90 & 1.79 & 1.30 & 4.02 & 2.66 \\
GPT-5.4-mini & 0.73 & 3.37 & 2.45 & 1.49 & 1.05 & 3.24 & 2.27 \\
Gemma4-E2B & 0.84 & 5.54 & 4.03 & 2.67 & 1.81 & 5.38 & 3.57 \\
InternVL & 1.34 & 10.40 & 7.89 & 5.58 & 4.27 & 10.63 & 7.56 \\
Gemini-2.5-Flash & 1.70 & 3.49 & 2.32 & 3.45 & 2.18 & 4.50 & 2.93 \\
PaddleOCR & 1.90 & 4.03 & 3.66 & 1.59 & 1.40 & 2.70 & 2.08 \\
DeepSeek & 2.04 & 5.95 & 4.35 & 3.83 & 2.42 & 4.97 & 3.22 \\
\bottomrule
\end{tabular}}
\end{table}

\begin{figure}[t!]
  \centering
  \includegraphics[width=0.9\columnwidth]{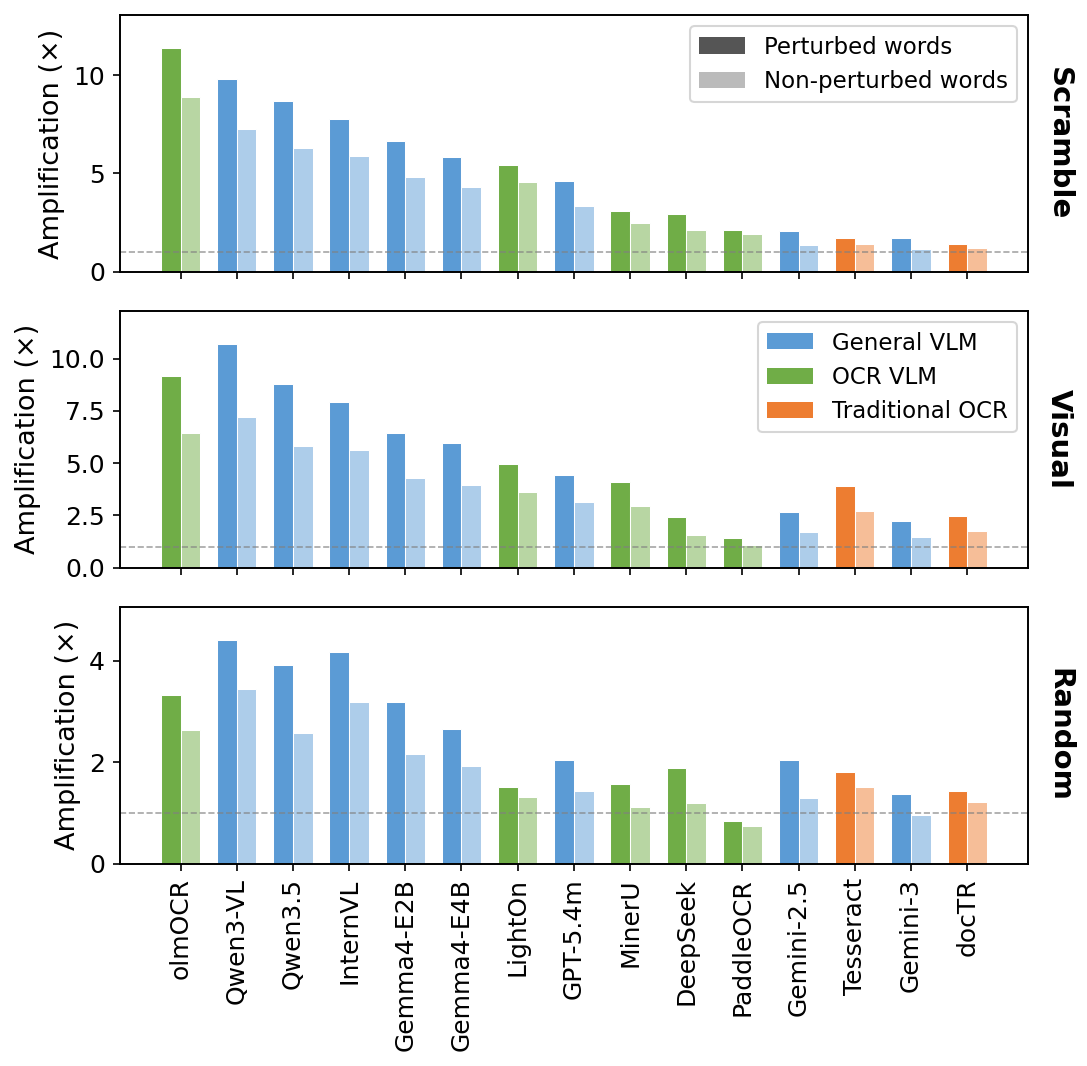}
  \caption{Error amplification on non-perturbed words. Perturbing ${\sim}5\%$ of characters induces substantial OCR errors on the remaining unperturbed text.}
  \label{fig:amplification}
\end{figure}

Perturbing ${\sim}5\%$ of characters also raises errors on the surrounding unperturbed text. Table~\ref{tab:collateral} reports per-word error rates on perturbed and non-perturbed positions, an attribution WER cannot provide since it aggregates over the whole document; this rate counts substitutions and deletions within each word group and is not directly comparable to WER. Figure~\ref{fig:amplification} plots the ratio against each model's clean original baseline. For general-purpose VLMs, error rates on non-perturbed words rise by up to 7.3$\times$ relative to baseline (Qwen3-VL-4B: 7.3$\times$ under scramble; Qwen3.5-4B: 6.3$\times$), while traditional OCR stays within 1.2--2.7$\times$ (Tesseract, docTR). The degree of propagation correlates with rewriting rates (Table~\ref{tab:error_types}): Qwen3.5-4B, which rewrites 27.4\% of perturbed words, shows 6.3$\times$ amplification on unperturbed words under scramble, while Gemini-3-Flash, which rewrites none, stays at 1.2$\times$. This suggests non-local errors and rewrites share an underlying mechanism rather than being independent failure modes.
This has two practical consequences. First, the cost of using a VLM for transcription cannot be estimated from the perturbation rate alone, since a small fraction of corrupted words contaminates a much larger fraction of the output. Second, error propagation is not predicted by clean-text performance: Gemini-3-Flash and Qwen3.5-4B have identical clean baselines (0.35\% per-word error rate for both), yet under scramble their amplification on unperturbed text differs more than fivefold (1.2$\times$ vs.\ 6.3$\times$).

\paragraph{Cross-lingual patterns.} As shown in Tables~\ref{tab:degradation_en}--\ref{tab:degradation_kr}, the three-tier ordering (traditional OCR $<$ OCR-specialized VLM $<$ general-purpose VLM) is true across English, Chinese, and Korean, so VLM rewriting is not specific to English. Which perturbation type damages each script the most differs. In English, scramble and visual perturbations both exceed random for nearly every system (14 of 15; Tesseract is the sole exception, at $+$0.17 vs.\ $+$0.20\,pp). In Chinese, the two available perturbations are closely matched: random and visual substitution differ by less than 0.5\,pp for 9 of the 14 systems, and the sign of the difference splits evenly (7 systems worse under random, 7 under visual). Random remains the harsher condition for the weakest readers (Gemma4-E2B 8.63 vs.\ 7.93\,pp; Gemma4-E4B 6.12 vs.\ 4.95\,pp; LightOnOCR-2-1B 3.46 vs.\ 2.30\,pp), but the gap closes for stronger ones (GPT-5.4-mini 2.26 vs.\ 2.32\,pp). Both halves of this pattern follow from the perturbation design: visual substitution is restricted to characters sharing radicals or stroke patterns, and random substitution draws replacements from the Level-1 common-character pool, so both conditions yield individually plausible characters rather than rare glyphs. In Korean, degradation is generally \textit{larger} than in English for the same model (GPT-5.4-mini averages $+$4.18\,pp vs.\ $+$1.50\,pp; Gemini-3-Flash $+$1.61\,pp vs.\ $+$0.12\,pp). This ordering depends on matching perturbation intensity across scripts. Korean words are short, so a fixed word-selection rate perturbs far fewer characters in Korean than in English and would make Korean appear the more robust language; calibrating every language to ${\sim}5\%$ character-level intensity (\S\ref{subsec:dataset}) removes this confound and reverses the apparent ordering.

\subsection{Uncovering the Rewriting Mechanism}
\label{subsec:probes}
The results in Table~\ref{tab:error_types} show that rewriting rates vary across VLMs, with the strongest-prior models rewriting up to 65\% of perturbed words. To understand \textit{where} this behavior originates, we probe the forward pass of Qwen3-VL-4B to locate the layers responsible for perturbation and rewriting. Qwen3-VL-4B has 36 transformer layers (L0–L35); we focus on every fourth layer.

\begin{figure}[t!]
  \centering
  \includegraphics[width=\columnwidth]{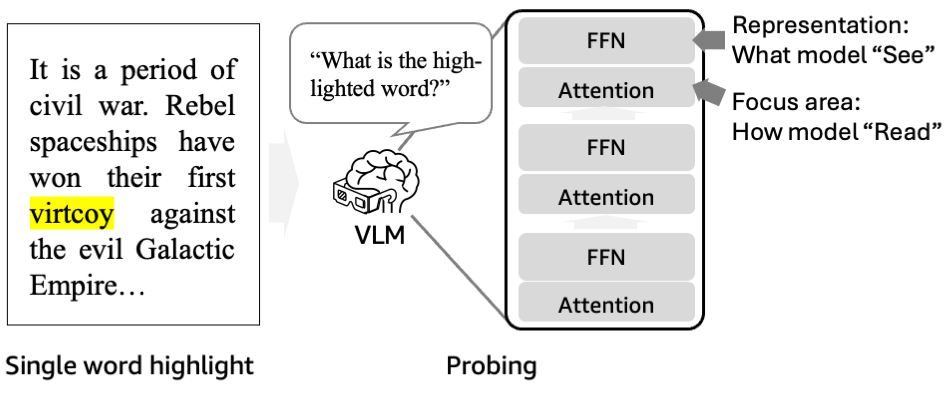}
  \caption{Probing setup. A single word is highlighted in yellow in the document image. The model is prompted to read the highlighted word.}
  \label{fig:probing-setup}
\end{figure}

\paragraph{Probe design.}
We sample one word uniformly at random from each of 500 English documents and apply the three perturbation types together with the unmodified original, yielding 2{,}000 (document, condition) pairs. The selected word is highlighted in yellow in the document image as shown in Figure~\ref{fig:probing-setup}, and the model is prompted to read the highlighted word.
  \begin{tcolorbox}[colback=gray!5, colframe=gray!50, boxrule=0.5pt, left=4pt, right=4pt, top=2pt, bottom=2pt]
\small\textbf{Probe Prompt:} ``What is the highlighted word in this document? Reply with only the word.''
\end{tcolorbox}

\begin{figure}[b!]
  \centering
  \includegraphics[width=\columnwidth]{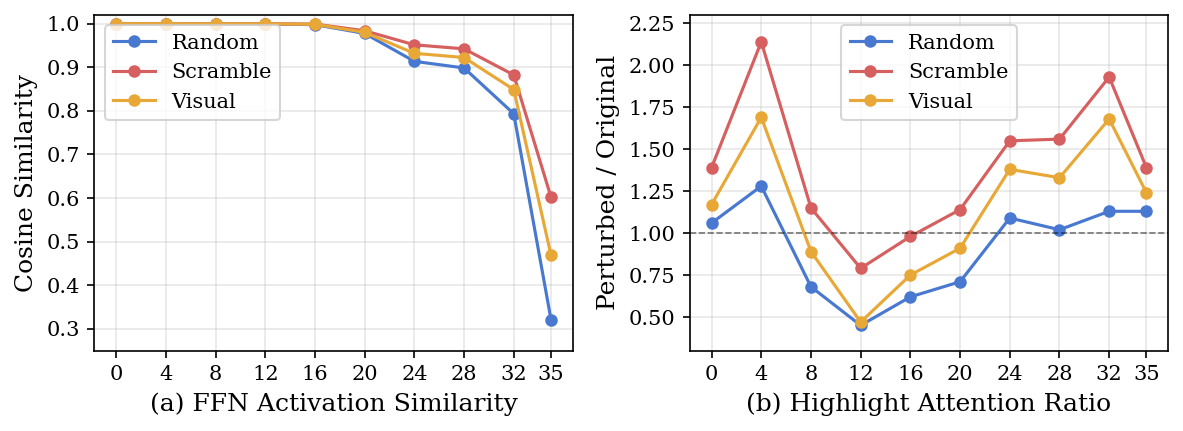}
  \caption{Per-layer FFN activation similarity and highlight attention ratio for Qwen3-VL-4B. Left: representations diverge monotonically. Right: attention to the perturbed highlight \emph{increases} (ratio $>$1) at early (L4) and late (L24--L35) layers.}
  \label{fig:internal_probes}
\end{figure}

For each \textit{(document, condition)} pair we record two layerwise quantities, the cosine similarity between FFN outputs at the last input token under the original and perturbed conditions, which approximates how much the perturbation has shifted the model's contextual representation before generation, and the attention weight on the highlighted word's image patches at the first generation step, which indicates where the model looks when it commits to its first output token.
\paragraph{Model detects perturbation, yet still rewrites.}
Table~\ref{tab:internal_probes_qwen3vl} and Figure~\ref{fig:internal_probes} report per-layer FFN similarity and attention ratios for all three perturbations. Representations are nearly identical through layer 12 (cosine similarity $>0.999$) and then diverge monotonically from layer 16 onward, reaching a minimum at layer 35 of 0.321 for random, 0.469 for visual, and 0.602 for scramble. This ordering, with random most divergent followed by visual and then scramble, matches how much each perturbation disrupts the word's lexical form. The model also allocates more attention to the perturbed word's image patches at both early (at L4, $2.14\times$ for scramble, $1.69\times$ for visual, and $1.28\times$ for random) and late (at L32, $1.93\times$ for scramble) layers, while mid-layers (L8--L16) show a dip. The model therefore re-reads the corrupted input, so rewriting is not a failure to notice the perturbation but a failure to act on it. The override must depend on something other than attention.
\begin{table}[b!]
\centering
\small
\caption{Per-layer FFN activation similarity and highlight attention ratio at first generation step. Attention ratio $>$1 = \emph{increased} attention to perturbed highlight.}
\label{tab:internal_probes_qwen3vl}
\resizebox{\columnwidth}{!}{%
\begin{tabular}{l|ccc|ccc}
\toprule
 & \multicolumn{3}{c|}{\textbf{FFN Activation Similarity}} & \multicolumn{3}{c}{\textbf{Highlight Attention Ratio}} \\
\cmidrule(lr){2-4} \cmidrule(lr){5-7}
\textbf{Layer} & \textbf{Random} & \textbf{Scramble} & \textbf{Visual} & \textbf{Random} & \textbf{Scramble} & \textbf{Visual} \\
\midrule
 0 & 0.9996 & 0.9999 & 0.9999 & 1.06$\times$ & 1.39$\times$ & 1.17$\times$ \\
 4 & 0.9995 & 0.9998 & 0.9998 & 1.28$\times$ & 2.14$\times$ & 1.69$\times$ \\
 8 & 0.9995 & 0.9998 & 0.9997 & 0.68$\times$ & 1.15$\times$ & 0.89$\times$ \\
12 & 0.9994 & 0.9998 & 0.9997 & 0.45$\times$ & 0.79$\times$ & 0.47$\times$ \\
16 & 0.9978 & 0.9991 & 0.9987 & 0.62$\times$ & 0.98$\times$ & 0.75$\times$ \\
20 & 0.9776 & 0.9838 & 0.9801 & 0.71$\times$ & 1.14$\times$ & 0.91$\times$ \\
24 & 0.9136 & 0.9515 & 0.9320 & 1.09$\times$ & 1.55$\times$ & 1.38$\times$ \\
28 & 0.8984 & 0.9423 & 0.9224 & 1.02$\times$ & 1.56$\times$ & 1.33$\times$ \\
32 & 0.7933 & 0.8824 & 0.8488 & 1.13$\times$ & 1.93$\times$ & 1.68$\times$ \\
35 & 0.3208 & 0.6019 & 0.4685 & 1.13$\times$ & 1.39$\times$ & 1.24$\times$ \\
\bottomrule
\end{tabular}}
\end{table}

\paragraph{Rewriting is gated by representation similarity.}
Tables~\ref{tab:probe_predictions_qwen3vl} and~\ref{tab:quartile_qwen3vl} report prediction outcomes and their relationship to FFN similarity. Rewriting occurs in 4.9\% of scramble cases and 6.0\% of visual cases but never under random substitution (0.0\%), since a random word shares no lexical overlap with any real word and leaves the language prior without a plausible target to recover. The relationship between representation shift and rewriting is sharper still when scramble samples are partitioned into quartiles by layer-35 FFN similarity. Samples in the bottom two quartiles, where the FFN representation has moved furthest from the original (similarity 0.34--0.45), are never rewritten, while samples in the top quartile, where the representation barely shifts (similarity $\geq$ 0.73), are rewritten at 4.66\%. Attention to the highlighted word does not track this pattern---if anything it moves the other way, falling slightly from Q1 to Q4 (0.00012 to 0.00009)---so what decides whether the model rewrites is not how hard it looks at the perturbed region but how far the perturbation has moved its internal representation. Rewriting happens precisely when the representation barely moves, leaving the language prior with a recognizable word to ``correct'' onto.

\begin{table}[t]
\centering
\small
\caption{Prediction breakdown on single-word probes. Model predicts original words with scramble (4.9\%) and visual (6\%) but not with random (0\%).}
\label{tab:probe_predictions_qwen3vl}
\resizebox{0.9\columnwidth}{!}{%
\begin{tabular}{lccccc}
\toprule
\textbf{Category} & \textbf{Original} & \textbf{Scramble} & \textbf{Random} & \textbf{Visual} \\
\midrule
Verbatim & 96.4\% & 69.2\% & 76.2\% & 62.8\% \\
Close & 0.6\% & 20.5\% & 17.2\% & 24.8\% \\
Moderate & 0.2\% & 1.5\% & 2.6\% & 2.6\% \\
Rewritten & --- & 4.9\% & 0.0\% & 6.0\% \\
Wrong & 2.8\% & 3.9\% & 4.0\% & 3.8\% \\
\bottomrule
\end{tabular}}
\end{table}
\begin{table}[t]
\centering
\small
\caption{Quartile analysis by FFN activation similarity at layer 35 (scramble perturbation). Q1 = most divergent, Q4 = most similar to original.}
\label{tab:quartile_qwen3vl}
\resizebox{\columnwidth}{!}{%
\begin{tabular}{lccc}
\toprule
\textbf{Quartile} & \textbf{FFN Sim.} & \textbf{Highlight Attn.} & \textbf{Rewritten} \\
\midrule
Q1 (most divergent) & 0.341 & 0.00012 & 0.0\% (0) \\
Q2                  & 0.454 & 0.00012 & 0.0\% (0) \\
Q3                  & 0.548 & 0.00011 & 0.24\% (1) \\
Q4 (most similar)   & 0.730 & 0.00009 & 4.66\% (19) \\
\bottomrule
\end{tabular}}
\end{table}

\begin{table}[!t]
\centering
\small
\caption{Prediction breakdown by word length (scramble). Shorter words are rewritten more frequently.}
\label{tab:word_length_qwen3vl}
\resizebox{\columnwidth}{!}{
\begin{tabular}{lcccccccc}
\toprule
\textbf{Length} & \textbf{n} & \textbf{Verbatim} & \textbf{Close} & \textbf{Moderate} & \textbf{Rewritten} & \textbf{Wrong} & \textbf{FFN (L35)} \\
\midrule
4 & 66 & 72.7\% & 16.7\% & 0.0\% & 7.6\% & 3.0\% & 0.607 \\
5 & 72 & 68.1\% & 18.1\% & 1.4\% & 8.3\% & 4.2\% & 0.584 \\
6 & 70 & 61.4\% & 15.7\% & 4.3\% & 10.0\% & 8.6\% & 0.578 \\
7 & 63 & 71.4\% & 20.6\% & 0.0\% & 3.2\% & 4.8\% & 0.475 \\
8--9 & 86 & 76.7\% & 20.9\% & 0.0\% & 0.0\% & 2.3\% & 0.458 \\
10+ & 52 & 61.5\% & 34.6\% & 3.8\% & 0.0\% & 0.0\% & 0.391 \\
\bottomrule
\end{tabular}}
\end{table}

\paragraph{``Wrong'' predictions reveal context leakage.}
The predictions classified as ``Wrong'' in Table~\ref{tab:probe_predictions_qwen3vl} are not random decoding errors. In most of these cases the model outputs a different salient word from the same document rather than a corrupted partial decoding, a behavior we call \textit{context leakage}. Context leakage and rewriting are two outcomes of the same underlying mechanism at different levels of representation disruption. Rewriting maps the perturbed word back to its original form, so the language prior still recognizes the corrupted token as a known word. Context leakage maps to an entirely different word, so the representation has been disrupted past the point where the prior can recover the original, and the model instead retrieves the most activated word from the surrounding document. Examples are provided in Appendix~\ref{sec:context_leakage}.
\paragraph{Word length affects rewriting rate.}
As shown in Table~\ref{tab:word_length_qwen3vl}, rewriting under scramble is concentrated in short words and absent for long ones. Words of 4--6 characters are rewritten 7.6--10.0\% of the time, peaking at 10.0\% for 6-character words. The rate then falls sharply at 7 characters (3.2\%) and reaches 0\% from 8 characters upward. Rewriting is not a smooth function of word length but a step function, with a clean cutoff between 7 and 8 characters where the language prior stops being able to recover the original form. Layer-35 similarity is 0.607 for 4-character words and falls to 0.391 for 10-character words, so shorter words shift the representation less and leave the language prior with a recognizable encoding to map back to the original. Longer words give the perturbation more characters to scramble, which produces enough cumulative disruption that the prior can no longer match the corrupted representation to a known word and the model falls back on what it sees.
\section{Conclusion}
\label{sec:conclusion}
In this work, we introduce \textit{FaithC4}, a controlled multilingual perturbation benchmark for measuring transcription faithfulness in VLMs, and use it to evaluate 15 systems across English, Chinese, and Korean. VLMs rewrite imperfect text into a more plausible form rather than transcribing it literally, a behavior that is negligible in traditional OCR systems, and the resulting three-tier ordering of degradation (traditional OCR $<$ OCR-specialized VLM $<$ general-purpose VLM) holds in all three languages. Probing Qwen3-VL-4B layer by layer shows that rewriting depends on how far the perturbation shifts the model's internal representation, not on how much attention the model allocates to the perturbed word. Rewriting occurs when the representation stays close to the clean encoding and disappears when the perturbation moves it far enough away, and the same representation-similarity mechanism explains why shorter words are rewritten more often than longer ones. Perturbing ${\sim}5\%$ of characters also increases errors on the remaining unperturbed text by up to 7.3$\times$, so a local perturbation has a non-local effect on document-level transcription. For applications that require literal transcription, such as legal document processing, historical manuscript digitization, and forensics, traditional OCR systems or carefully selected OCR-specialized VLMs remain the safest choice, while general-purpose VLMs introduce systematic rewriting that clean-text benchmarks do not capture. We release \textit{FaithC4}, together with our perturbation and evaluation code, at \url{https://github.com/gwanglee/DoVLMsRead}.

\section{Limitations}
\label{sec:limitations}

Our study has several limitations. First, while we evaluate three languages (English, Korean, Chinese), behavior may differ for other scripts (e.g., Arabic, Devanagari) or specialized domains (e.g., mathematical notation, code). Second, our perturbation types---scramble, random, and visual---represent controlled manipulations rather than naturally occurring text degradations such as scanning artifacts, handwriting variation, or low-resolution capture. Third, our dataset uses synthetically rendered PDFs, which may not capture the visual complexity of real scanned documents. Fourth, while the full-document evaluation covers all 15 models, our mechanistic probing focuses on a single model (Qwen3-VL); the internal dynamics may differ across architectures. Fifth, we report two mitigation experiments in Appendix~\ref{sec:ablations} (no-correction prompt; chain-of-thought reasoning), but do not investigate fine-tuning, decoding-time interventions, or specialized OCR-faithfulness training. Finally, the traditional OCR baselines (Tesseract, docTR) have limited multilingual support, restricting cross-category comparison for Korean and Chinese.


\bibliography{anthology}

\clearpage

\appendix

\section*{Appendix}
\label{sec:appendix}

\renewcommand{\thetable}{A.\arabic{table}}
\setcounter{table}{0}
\renewcommand{\thefigure}{A.\arabic{figure}}
\setcounter{figure}{0}

\section{Details for Perturbation Generation}
\label{sec:cjk_perturbation}

While the three perturbation types (scramble, random, visual) are straightforward for English (Latin alphabet, whitespace-delimited words), their application to Chinese and Korean requires language-specific adaptations.

\paragraph{Word segmentation.}
English words are delimited by whitespace. Chinese lacks word boundaries entirely; we define ``words'' as non-overlapping 4-character windows, yielding approximately 250 windows per document. Korean text is also whitespace-delimited, but many tokens are short function particles (1--3 characters) that do not meet the minimum length threshold.

\paragraph{Eligibility filtering.}
A word is eligible for perturbation if it contains $\geq$4 characters and consists entirely of script-appropriate characters (ASCII letters for English, Hangul syllables for Korean, CJK Unified Ideographs for Chinese). This filter excludes punctuation, numbers, mixed-script tokens, and short function words. Rather than fixing the selection rate, we calibrate the number of units perturbed separately for each language and condition so that all of them change ${\sim}5\%$ of the characters in a document:

\begin{itemize}
    \item \textbf{English}: 24.6--46.1 words/doc perturbed (4.1--7.7\% of words), changing 5.08--5.20\% of characters
    \item \textbf{Chinese}: 24.5--46.0 windows/doc perturbed (9.8--18.4\% of windows), changing 4.98--5.03\% of characters
    \item \textbf{Korean}: 17.4--30.8 words/doc perturbed (3.9--6.9\% of words), changing 4.65--5.04\% of characters
\end{itemize}

Chinese requires the largest share of units because its 4-character windows are shorter than the average English (5.9 characters) or Korean (5.0 characters) word, so each perturbed window contributes fewer changed characters. Holding the character-level intensity fixed at ${\sim}5\%$ means cross-lingual comparisons in \S\ref{sec:results} are not confounded by differing perturbation strength, which was a limitation of earlier versions of this benchmark.

\paragraph{Random.}
Each character in a selected word is replaced with a different character from the same language. For Chinese, each character is replaced with another Chinese character drawn from the Level-1 pool of the Table of General Standard Chinese Characters ($3{,}500$ common characters), so that replacements are individually plausible glyphs rather than rare ones. For Korean, each syllable character is replaced with another from the 11,172 syllables that exist and are commonly used in the language. In all three languages, the result is a sequence of individually valid characters that together form no recognizable word.

\paragraph{Scramble.}
For English and Korean, characters within each selected word are randomly permuted (keeping first and last characters fixed for words $\geq$5 characters). Scramble is not applied to Chinese, which lacks space-delimited word boundaries and therefore has no word-level unit within which to apply character reordering.

\paragraph{Visual.}
Each character is replaced with a different character that looks similar to the original. For English, we use a manually curated confusion map where character shapes overlap. Table~\ref{tab:visual_map} shows representative mappings. Note that some substitutions produce multi-character sequences (e.g., \texttt{m}$\to$\texttt{rn}, \texttt{d}$\to$\texttt{cl}) that are visually indistinguishable at typical reading sizes.

For Chinese, each character is replaced with another that shares visual components --- for example, characters built from the same radical or with similar stroke patterns. We use the top-2,000 most frequent characters and their visually similar counterparts (e.g., \begin{CJK}{UTF8}{gbsn}大$\to$太, 人$\to$入\end{CJK}). For Korean, each syllable is decomposed into its consonant and vowel components, and one component is swapped with a visually similar one, producing a syllable that looks close to the original but represents a different sound. In all cases, the substitution is a real, displayable character --- the perturbation is designed to test whether the model reads what is actually on the page or infers what \emph{should} be there based on context.


\begin{table}[h!]
\centering
\small
\caption{English visual confusion map. Each source character maps to one or more visually similar alternatives.}
\label{tab:visual_map}
\resizebox{0.75\columnwidth}{!}{
\begin{tabular}{cl|cl}
\toprule
\textbf{Char} & \textbf{Confusables} & \textbf{Char} & \textbf{Confusables} \\
\midrule
a & e, o, c & A & H, N \\
b & h, d & B & S, R, D, E \\
c & e, o & C & G, O \\
d & o, cl, b & D & O, B \\
e & a, c, o & E & F, B \\
f & l & F & E \\
g & q, c, o & G & C, O, Q \\
h & n, b, li & H & A, N \\
i & l, j & I & l, J, T \\
j & i, l & J & I, T \\
k & lc, l, x & K & X, Y \\
l & I, i, j, f & L & I, N \\
m & n, rn & M & N, H \\
n & h, m & N & M, H \\
o & a, e, c & O & C, D, G, Q \\
p & q, o & P & R, B \\
q & g, p & Q & O, G, C \\
r & n, c & R & P, B \\
s & a, e, z & S & B \\
t & i, l & T & I, J \\
u & v, n & U & V \\
v & u & V & U, N \\
w & vv, u, m & W & V, U, M, VV \\
x & k & X & K \\
y & j & Y & I, K \\
z & s, e & Z & S \\
\bottomrule
\end{tabular}}
\end{table}

\section{Models}
\label{sec:app_models}

Table~\ref{tab:models} provides inference configuration details for all evaluated models.

\begin{table}[h!]
\centering
\caption{Models evaluated. docTR is evaluated on English only.}
\label{tab:models}
\resizebox{\columnwidth}{!}{
\begin{tabular}{llcl}
\toprule
\textbf{Category} & \textbf{Model} & \textbf{Size} & \textbf{Prompt} \\
\midrule
\multirow{8}{*}{\shortstack[l]{General\\VLM}}
& Gemini-3-Flash & --- & \multirow{8}{*}{\textit{\shortstack[l]{Extract all text from this\\document image. Output\\document text only.}}} \\
& Gemini-2.5-Flash & --- & \\
& GPT-5.4-mini & --- & \\
& Qwen3.5-4B & 4B & \\
& Qwen3-VL-4B & 4B & \\
& InternVL3.5-4B & 4B & \\
& Gemma4-E4B & 4B & \\
& Gemma4-E2B & 2B & \\
\midrule
\multirow{5}{*}{\shortstack[l]{OCR-Specialized\\VLM}}
& olmOCR-2-7B & 7B & olmocr library \\
& DeepSeek-OCR-2 & 2B & \texttt{<|grounding|>} \\
& MinerU2.5-Pro & 1.2B & N/A (two-step) \\
& PaddleOCR-VL-1.5 & 1.5B & \texttt{OCR:} \\
& LightOnOCR-2-1B & 1B & N/A (image-only) \\
\midrule
\multirow{2}{*}{\shortstack[l]{Traditional\\OCR}}
& Tesseract & --- & N/A \\
& docTR & --- & N/A \\
\bottomrule
\end{tabular}}
\end{table}

\section{Failure Detection and Exclusion}
\label{app:failures}

Before computing OCR metrics, we apply a rule-based filter to identify model outputs that represent complete failures rather than OCR errors. A document is excluded from evaluation if any of the following conditions are met:

\begin{itemize}
\item \textbf{Empty output}: fewer than 20 characters produced.
\item \textbf{Refusal}: output begins with phrases indicating task refusal (e.g., ``I'm sorry'', ``I cannot'').
\item \textbf{Hex dump}: $>$70\% of alphanumeric characters are hexadecimal digits, indicating the model output Unicode codepoints as ASCII rather than rendering text.
\item \textbf{Repetition}: vocabulary diversity $<$5\% (fewer than 5 unique tokens per 100), indicating degenerate repetition loops.
\item \textbf{Script mismatch}: for Korean/Chinese documents, the output contains $<$1\% target-script characters when the ground truth contains $>$10\%, indicating the model produced text in the wrong writing system.
\end{itemize}

\paragraph{Failure rate analysis.}

Table~\ref{tab:failure} and \ref{tab:failure_by_pert} shows failure counts by model and language (summed across all four perturbation conditions). Two key observations are:

\textit{First, failures are language-specific, not uniform.} Almost all models achieve very low failure counts on English (at most 4 documents out of 2{,}000; PaddleOCR-VL-1.5 is the sole exception at 48), but certain model--language pairs exhibit high failure rates: Gemma4-E2B fails on 444 Chinese documents (29.6\%), PaddleOCR-VL-1.5 on 260 Korean documents (14.3\%), InternVL3.5-4B on 144 Chinese documents (9.6\%), and Qwen3-VL-4B on 119 Korean documents (6.5\%). These represent fundamental model limitations for specific languages.

\textit{Second, most failures are perturbation-independent.} Table~\ref{tab:failure_by_pert} shows failure counts by perturbation type (aggregated over languages). For most models, failures are roughly uniform across original, scramble, random, and visual conditions---confirming that failures reflect baseline model limitations rather than perturbation-induced errors. The apparently lower scramble counts reflect coverage, not robustness: Chinese has no scramble condition, so the scramble column is computed over English and Korean only (955 documents) while the other three cover all three languages (1{,}455 documents). Restricted to English and Korean, scramble accounts for 0.93--1.05 times the uniform one-quarter share for the three models with the most failures (Qwen3-VL-4B, PaddleOCR-VL-1.5, Gemma4-E2B).

\begin{table}[t]
\centering
\caption{Document failure counts by language (summed across perturbation conditions). Failures are excluded from OCR metrics. Content failures only (refusal/repetition/hallucination/truncation/empty).}
\label{tab:failure}
\resizebox{\columnwidth}{!}{
\begin{tabular}{lcccc}
\toprule
\textbf{Model} & \textbf{EN} & \textbf{ZH} & \textbf{KO} & \textbf{Dominant type} \\
\midrule
docTR & 0 & --- & --- & -- \\
Gemini-3-Flash & 0 & 8 (0.5\%) & 0 & repetition \\
GPT-5.4-mini & 4 (0.2\%) & 5 (0.3\%) & 2 (0.1\%) & truncation \\
olmOCR-2-7B & 0 & 17 (1.1\%) & 14 (0.8\%) & repetition \\
Tesseract & 0 & 31 (2.1\%) & 0 & hallucination \\
DeepSeek-OCR-2 & 1 (0.1\%) & 16 (1.1\%) & 23 (1.3\%) & repetition \\
LightOnOCR-2-1B & 0 & 29 (1.9\%) & 14 (0.8\%) & repetition \\
Qwen3.5-4B & 2 (0.1\%) & 24 (1.6\%) & 44 (2.4\%) & hallucination \\
MinerU2.5-Pro & 0 & 49 (3.3\%) & 21 (1.2\%) & hallucination \\
Gemini-2.5-Flash & 1 (0.1\%) & 91 (6.1\%) & 22 (1.2\%) & hallucination \\
Gemma4-E4B & 0 & 108 (7.2\%) & 7 (0.4\%) & repetition \\
InternVL3.5-4B & 2 (0.1\%) & 144 (9.6\%) & 24 (1.3\%) & hallucination \\
Qwen3-VL-4B & 1 (0.1\%) & 75 (5.0\%) & 119 (6.5\%) & hallucination \\
PaddleOCR-VL-1.5 & 48 (2.4\%) & 87 (5.8\%) & 260 (14.3\%) & repetition \\
Gemma4-E2B & 1 (0.1\%) & 444 (29.6\%) & 128 (7.0\%) & repetition \\
\bottomrule
\end{tabular}}
\end{table}


\begin{table}[t]
\centering
\caption{Document failure counts by perturbation type (aggregated over EN/ZH/KO). Sorted by total failures. Content failures only.}
\label{tab:failure_by_pert}
\resizebox{\columnwidth}{!}{
\begin{tabular}{lcccc}
\toprule
\textbf{Model} & \textbf{Original} & \textbf{Scramble} & \textbf{Random} & \textbf{Visual} \\
\midrule
docTR & 0 & 0 & 0 & 0 \\
Gemini-3-Flash & 6 (0.4\%) & 0 & 1 (0.1\%) & 1 (0.1\%) \\
GPT-5.4-mini & 6 (0.4\%) & 2 (0.2\%) & 2 (0.1\%) & 1 (0.1\%) \\
olmOCR-2-7B & 9 (0.6\%) & 2 (0.2\%) & 10 (0.7\%) & 10 (0.7\%) \\
Tesseract & 9 (0.6\%) & 0 & 12 (0.8\%) & 10 (0.7\%) \\
DeepSeek-OCR-2 & 11 (0.8\%) & 6 (0.6\%) & 12 (0.8\%) & 11 (0.8\%) \\
LightOnOCR-2-1B & 15 (1.0\%) & 4 (0.4\%) & 13 (0.9\%) & 11 (0.8\%) \\
Qwen3.5-4B & 16 (1.1\%) & 9 (0.9\%) & 22 (1.5\%) & 23 (1.6\%) \\
MinerU2.5-Pro & 22 (1.5\%) & 3 (0.3\%) & 21 (1.4\%) & 24 (1.6\%) \\
Gemini-2.5-Flash & 33 (2.3\%) & 4 (0.4\%) & 38 (2.6\%) & 39 (2.7\%) \\
Gemma4-E4B & 35 (2.4\%) & 2 (0.2\%) & 53 (3.6\%) & 25 (1.7\%) \\
InternVL3.5-4B & 51 (3.5\%) & 6 (0.6\%) & 62 (4.3\%) & 51 (3.5\%) \\
Qwen3-VL-4B & 51 (3.5\%) & 28 (2.9\%) & 58 (4.0\%) & 58 (4.0\%) \\
PaddleOCR-VL-1.5 & 108 (7.4\%) & 76 (8.0\%) & 120 (8.2\%) & 91 (6.3\%) \\
Gemma4-E2B & 179 (12.3\%) & 34 (3.6\%) & 173 (11.9\%) & 187 (12.9\%) \\
\bottomrule
\end{tabular}}
\end{table}


\paragraph{Intersection-based evaluation.} Our evaluation further ensures fairness: for each (model, language) pair, we compute metrics only on documents that the model processed successfully across \emph{all four} perturbation conditions. This eliminates survivorship bias from differential failure rates across perturbation types.

\section{Baseline Performance}
\label{appendix:baseline}

Table~\ref{tab:baselines} reports the OCR performance of all models on original (unperturbed) documents. These baselines establish each model's reading ability independent of perturbation effects, and serve as the reference point for computing degradation ($\Delta$) values reported in the main text.

Models are grouped by category: general-purpose VLMs, OCR-specialized VLMs, and traditional OCR engines. Within each group, models are sorted by English WER. Key observations are:

\begin{itemize}
    \item General VLMs achieve strong English baselines, with seven of eight below 1.5\% WER; the best (Qwen3.5-4B at 0.22\%) approaches near-perfect transcription.
    \item OCR-specialized models achieve competitive baseline accuracy at significantly smaller parameter counts. LightOnOCR-2-1B (0.14\%) and MinerU2.5-Pro (0.16\%) beat every general VLM on English despite their far smaller size, and olmOCR-2-7B leads all systems on Korean (4.03\% WER). On Chinese, olmOCR-2-7B is the strongest OCR-specialized system (1.94\% CER) but is edged out by Gemini-3-Flash (0.77\%) and Qwen3.5-4B (1.37\%). This suggests that task-specific training largely compensates for reduced model capacity on OCR tasks.
    \item Traditional OCR is competitive on English (Tesseract 0.31\%, docTR 0.55\%) but collapses on Korean: Tesseract reaches 40.10\% WER, and docTR supports neither Korean nor Chinese. This reflects the limited multilingual coverage of pre-neural OCR pipelines.
    \item Both CJK languages are far harder than English for every system, but their relative difficulty is metric-dependent: Chinese CER is lower than Korean WER for 12 of 14 systems, yet on EDS---the only metric computed identically across all three languages---Korean scores higher than Chinese for 11 of 14. We therefore do not claim an ordering between the two. What is robust is the spread: English baselines fall in a narrow 0.14--3.04\% WER band, whereas Chinese and Korean EDS span 23 and 27 points respectively, so the CJK settings are where systems actually separate.
\end{itemize}

\begin{table}[ht]
\centering
\caption{Baseline (original, unperturbed) performance for all models. ``N/A'' indicates the model does not support that language.}
\label{tab:baselines}
\resizebox{\columnwidth}{!}{
\begin{tabular}{l cc cc cc}
\toprule
& \multicolumn{2}{c}{\textbf{EN}} & \multicolumn{2}{c}{\textbf{ZH}} & \multicolumn{2}{c}{\textbf{KO}} \\
\cmidrule(lr){2-3} \cmidrule(lr){4-5} \cmidrule(lr){6-7}
\textbf{Model} & WER$\downarrow$ & EDS$\uparrow$ & CER$\downarrow$ & EDS$\uparrow$ & WER$\downarrow$ & EDS$\uparrow$ \\
\midrule
\multicolumn{7}{l}{\textit{General VLM}} \\
Qwen3.5-4B         & 0.22 & 99.95 & 1.37 & 97.92 & 6.46 & 98.10 \\
Qwen3-VL-4B        & 0.30 & 99.92 & 2.47 & 97.67 & 6.99 & 96.85 \\
Gemini-3-Flash     & 0.34 & 99.94 & 0.77 & 98.04 & 4.85 & 99.05 \\
GPT-5.4-mini       & 0.40 & 99.91 & 3.80 & 94.87 & 8.23 & 98.05 \\
Gemma4-E4B         & 0.67 & 99.85 & 13.03 & 88.46 & 12.84 & 95.88 \\
Gemma4-E2B         & 0.87 & 99.75 & 35.45 & 75.02 & 19.73 & 91.62 \\
InternVL3.5-4B     & 1.38 & 99.57 & 4.82 & 93.94 & 18.60 & 95.39 \\
Gemini-2.5-Flash   & 2.46 & 99.49 & 5.88 & 92.97 & 7.84 & 98.02 \\
\midrule
\multicolumn{7}{l}{\textit{OCR-Specialized VLM}} \\
LightOnOCR-2-1B    & 0.14 & 99.96 & 6.23 & 94.22 & 8.26 & 97.38 \\
MinerU2.5-Pro      & 0.16 & 99.93 & 3.44 & 96.46 & 5.27 & 98.23 \\
olmOCR-2-7B        & 0.25 & 99.85 & 1.94 & 97.98 & 4.03 & 98.59 \\
PaddleOCR-VL-1.5   & 1.48 & 99.32 & 7.17 & 94.29 & 20.55 & 87.01 \\
DeepSeek-OCR-2     & 3.04 & 99.52 & 2.80 & 91.10 & 35.49 & 91.19 \\
\midrule
\multicolumn{7}{l}{\textit{Traditional OCR}} \\
Tesseract          & 0.31 & 99.90 & 5.53 & 93.35 & 40.10 & 72.26 \\
docTR              & 0.55 & 99.85 & N/A & N/A & N/A & N/A \\
\bottomrule
\end{tabular}}
\end{table}

\section{Context Leakage in ``Wrong'' Predictions}
\label{sec:context_leakage}

When the highlighted word is heavily perturbed, the model occasionally outputs a completely unrelated word---neither the perturbed text (verbatim) nor the original word (rewritten). These ``Wrong'' predictions account for 3.8--4.0\% of cases across perturbation types (Table~\ref{tab:probe_predictions_qwen3vl}). Analysis reveals they are predominantly \emph{context leakage}: the model reads a salient word from elsewhere in the document instead of the highlighted target.

Table~\ref{tab:context_leakage_examples} shows representative cases where the model produces the same wrong prediction across all three perturbation types for the same document. The predicted word is always a topically prominent term from the document (e.g., \textit{innovation} from an article about disruptive innovation, \textit{keto} from a health audiobook listing). The perturbation-invariance (identical wrong answer regardless of perturbation type) rules out partial decoding as the cause. Instead, when the highlighted word becomes unreadable, the model defaults to the most contextually salient word in the document. The large majority of ``Wrong'' predictions across the three perturbation types are context leakage of this kind.

\begin{table}[h]
\centering
\small
\caption{Context leakage examples. The model produces the same wrong prediction across all perturbation types, indicating document-level saliency rather than partial decoding.}
\label{tab:context_leakage_examples}
\resizebox{\columnwidth}{!}{%
\begin{tabular}{llll}
\toprule
\textbf{Original Word} & \textbf{Predicted (all perts)} & \textbf{Document Topic} \\
\midrule
Leonardo   & innovation  & Article on disruptive innovation \\
health     & keto        & Audiobook listing (keto diet) \\
coverage   & robocalls   & News article on robocalls \\
being      & hdbaset     & Forum post on HDBaseT technology \\
Meryl      & likability  & Article on leadership likability \\
moving     & tech        & Article on tech industry \\
\bottomrule
\end{tabular}}
\end{table}

\section{Ablation Studies}
\label{sec:ablations}

\subsection{Effect of model size.}

To examine whether perturbation sensitivity scales with model capacity, we evaluate two model families across multiple sizes on English: Qwen3.5 (0.8B, 2B, 4B, 9B) and Qwen3-VL (2B, 4B, 8B, 30B). All variants use the same prompt and inference configuration, differing only in parameter count. Table~\ref{tab:size_scaling} reports the results.

\begin{table}[t]
\centering
\caption{Perturbation degradation by model size (English). Baseline improves with size, but degradation plateaus beyond 4B.}
\label{tab:size_scaling}
\resizebox{\columnwidth}{!}{
\begin{tabular}{llcccccccc}
\toprule
& & \multicolumn{4}{c}{$\Delta$WER (pp$\downarrow$)} & \multicolumn{4}{c}{$\Delta$EDS (pp$\uparrow$)} \\
\cmidrule(lr){3-6} \cmidrule(lr){7-10}
\textbf{Family} & \textbf{Size} & Ori & Scr & Vis & Ran & Ori & Scr & Vis & Ran \\
\midrule
Qwen3.5 & 0.8B & 0.39 & +1.80 & +1.93 & +1.15 & 99.88 & $-$0.49 & $-$0.41 & $-$0.25 \\
Qwen3.5 & 2B   & 0.42 & +0.97 & +1.00 & +0.47 & 99.81 & $-$0.20 & $-$0.12 & +0.03 \\
Qwen3.5 & 4B   & 0.25 & +1.48 & +1.63 & +0.74 & 99.94 & $-$0.43 & $-$0.41 & $-$0.10 \\
Qwen3.5 & 9B   & 0.14 & +1.64 & +1.44 & +0.81 & 99.96 & $-$0.62 & $-$0.41 & $-$0.20 \\
Qwen3.5 & 27B  & 0.17 & +1.44 & +1.35 & +0.78 & 99.96 & $-$0.53 & $-$0.40 & $-$0.22 \\
\midrule
Qwen3-VL & 2B  & 0.58 & +1.04 & +1.27 & +0.53 & 99.79 & $-$0.20 & $-$0.31 & $-$0.03 \\
Qwen3-VL & 4B  & 0.29 & +2.04 & +2.05 & +1.69 & 99.92 & $-$0.60 & $-$0.51 & $-$0.38 \\
Qwen3-VL & 8B  & 0.17 & +1.52 & +1.50 & +1.08 & 99.95 & $-$0.48 & $-$0.39 & $-$0.15 \\
Qwen3-VL & 30B & 0.21 & +1.88 & +1.55 & +0.72 & 99.94 & $-$0.71 & $-$0.41 & $-$0.12 \\
\bottomrule
\end{tabular}}
\end{table}

Baseline OCR quality improves monotonically with scale: Qwen3.5 drops from 0.39\% WER at 0.8B to 0.14\% at 9B. However, perturbation sensitivity is non-monotonic and plateaus beyond 4B. In both families, the 2B variant exhibits the lowest average degradation (+0.81pp and +0.94pp), while the 4B models show the highest (+1.28pp and +1.92pp). Scaling further to 8B--30B does not increase degradation: Qwen3-VL-8B (+1.37pp) and Qwen3-VL-30B (+1.38pp) are both lower than the 4B variant. This non-monotonic pattern (lower degradation at 2B, peak at 4B, plateau beyond) may reflect competing effects of language capacity and visual encoding quality, but disentangling these factors requires further investigation.

\subsection{Effect of explicit instruction.}
We test whether instructing the model not to correct errors reduces rewriting behavior by appending an explicit no-correction clause to the base prompt.

\begin{tcolorbox}[colback=gray!5, colframe=gray!50, boxrule=0.5pt, left=4pt, right=4pt, top=2pt, bottom=2pt]
\small\textbf{No-correction prompt:} ``Extract all text from this document image. This document may contain typos, scrambled words, or random character sequences. Transcribe everything exactly as it appears --- do not correct or fix any words. Output document text only.''
\end{tcolorbox}

Table~\ref{tab:prompt_ablation} shows results for English on two models. For Qwen3-VL-4B, the no-correction instruction reduces degradation WER for all perturbation types (scramble: 2.04$\to$1.07\%, visual: 2.05$\to$1.33\% and random: 1.69$\to$ 1.13\%). 
For Qwen3.5-4B, reductions are smaller but also consistent (scramble: 1.48$\to$1.04\%, visual 1.63$\to$1.10\%, random: 0.74$\to$0.50\%). The instruction does not eliminate degradation but reduces its magnitude by roughly 30--50\% across perturbation types. One side effect is a slight increase in baseline WER ( 0.29$\to$0.66\% for Qwen3-VL, 0.25$\to$0.34\% for Qwen3.5), indicating the model becomes more conservative in its transcription when warned about potential errors.

\subsection{Effect of chain-of-thought reasoning.}
We additionally test whether enabling thinking helps the model follow the no-correction instruction. The ``no\_correct + think'' rows in Table~\ref{tab:prompt_ablation} show minimal additional benefit: for Qwen3-VL, scramble degradation increases slightly from $+$1.07 to $+$1.21\,pp; for Qwen3.5, it decreases slightly from $+$1.04 to $+$1.07\,pp (scramble) but improves on random ($+$0.50$\to$$+$0.41\,pp). Neither model generates substantive reasoning tokens (Qwen3-VL produces no thinking tokens, and Qwen3.5 generates empty \texttt{<think></think>} blocks) confirming that these models treat OCR as a direct perception task that does not benefit from deliberative reasoning.

\begin{table}[h!]
\centering
\caption{Prompt ablation (English). \textbf{Bold}: improved vs.\ baseline; \underline{underlined}: degraded.}
\label{tab:prompt_ablation}
\resizebox{\columnwidth}{!}{
\begin{tabular}{llcccccccc}
\toprule
& & \multicolumn{4}{c}{$\Delta$WER (pp$\downarrow$)} & \multicolumn{4}{c}{$\Delta$EDS (pp$\uparrow$)} \\
\cmidrule(lr){3-6} \cmidrule(lr){7-10}
\textbf{Model} & \textbf{Condition} & Ori & Scr & Vis & Ran & Ori & Scr & Vis & Ran \\
\midrule
Qwen3-VL-4B & baseline      & 0.29 & +2.04 & +2.05 & +1.69 & 99.92 & $-$0.60 & $-$0.51 & $-$0.38 \\
Qwen3-VL-4B & no\_correct   & \underline{0.66} & \textbf{+1.07} & \textbf{+1.33} & \textbf{+1.13} & \underline{99.78} & \textbf{$-$0.11} & \textbf{$-$0.16} & \textbf{$-$0.06} \\
Qwen3-VL-4B & no\_correct+think & 0.29 & \textbf{+1.21} & \textbf{+1.52} & \textbf{+1.26} & \textbf{99.93} & \textbf{$-$0.22} & \textbf{$-$0.27} & \textbf{$-$0.18} \\
\midrule
Qwen3.5-4B  & baseline      & 0.25 & +1.48 & +1.63 & +0.74 & 99.94 & $-$0.43 & $-$0.41 & $-$0.10 \\
Qwen3.5-4B  & no\_correct   & \underline{0.34} & \textbf{+1.04} & \textbf{+1.10} & \textbf{+0.50} & 99.94 & \textbf{$-$0.27} & \textbf{$-$0.29} & \textbf{$-$0.08} \\
Qwen3.5-4B  & no\_correct+think & \underline{0.44} & \textbf{+1.07} & \textbf{+0.98} & \textbf{+0.41} & \underline{99.78} & \textbf{$-$0.16} & \textbf{$-$0.16} & \textbf{$+$0.08} \\
\bottomrule
\end{tabular}}
\end{table}
\section{Validation on Real-World Documents}
\label{appendix:downstream}

To validate that FaithC4's findings extend to real-world documents, we evaluate ASIN extraction on 150 PDF documents from the product compliance domain.

\paragraph{Task.} Models are asked to identify the ASINs in each document (10-character alphanumeric product identifiers. e.g., \texttt{B0CNSR37BD}). ASINs are random strings carrying no semantic content, functionally equivalent to FaithC4's \textit{random} perturbation condition. FaithC4 shows that random perturbation produces 0\% rewrite rate in single-word probes (Table~\ref{tab:probe_predictions_qwen3vl}) because the language prior has no correction target, yet still degrades document-level WER by up to 3.2\,pp through non-local disruption (\S\ref{subsec:degradation}). We test whether naturally occurring random strings exhibit the same pattern.

\paragraph{Setup.} We compare (1)~\textbf{Direct VLM}: page images passed to the model with an extraction prompt, and (2)~\textbf{OCR$\rightarrow$LLM pipeline}: pages transcribed by OCR, then text passed to Claude Sonnet 4.5 for extraction. ASIN extraction is measured by set-based F1 (exact match).

\begin{table}[h]
\centering
\small
\caption{ASIN extraction F1 on 150 real documents, ordered by F1. \textit{Direct} passes page images to the model; \textit{Pipeline} transcribes with the listed OCR model and passes the text to LLM for extraction.}
\label{tab:downstream_asin}
\begin{tabular}{lcc}
\toprule
\textbf{Model} & \textbf{Setting} & \textbf{ASIN F1} \\
\midrule
MinerU2.5-Pro    & Pipeline & 99.11\% \\
LightOnOCR-2-1B  & Pipeline & 98.36\% \\
docTR            & Pipeline & 96.08\% \\
Qwen3.5-4B       & Direct   & 93.79\% \\
olmOCR-2-7B      & Pipeline & 93.48\% \\
DeepSeek-OCR-2   & Pipeline & 91.90\% \\
Tesseract        & Pipeline & 86.39\% \\
Gemma4-E2B       & Direct   & 77.58\% \\
PaddleOCR-VL-1.5 & Pipeline & 74.48\% \\
Gemma4-E4B       & Direct   & 68.64\% \\
InternVL3.5-4B   & Direct   & 47.45\% \\
Qwen3-VL-4B      & Direct   & 35.25\% \\
\bottomrule
\end{tabular}
\end{table}

\paragraph{Analysis.} Pipelines outperform direct VLMs by a wide margin (median 93.5\% vs.\ 68.6\% F1): every pipeline except PaddleOCR-VL-1.5 beats every direct VLM except Qwen3.5-4B. Pipelines separate reading from understanding, transcribing character sequences regardless of semantic content; consistent with this, both traditional OCR backends land in the upper half of the pipeline group (docTR 96.1\%, Tesseract 86.4\%) despite having no language model to exploit.

Measured perturbation sensitivity predicts failure better than category membership does. Across the 12 systems evaluated in both settings, English degradation under the \textit{random} condition---the FaithC4 analog of an ASIN---correlates strongly with ASIN F1 (Spearman $\rho=-0.76$, $p=0.005$). InternVL3.5-4B has the largest random-condition degradation (3.21\,pp) and the second-lowest F1 (47.5\%), and the two systems that cross the category boundary are the ones FaithC4 flags: robust Qwen3.5-4B ($\leq$2.29\,pp) leads the direct models, while sensitive PaddleOCR-VL-1.5 (3.19\,pp) trails the pipelines. Qwen3-VL-4B is the one clear miss, with mid-range document-level degradation (1.33\,pp) but the lowest F1 overall (35.3\%).


\end{document}